# Combining Uncertain Estimates

*Henry Hamburger*

George Mason University

and

Naval Research Laboratory

## 1. Abstract = Overview

In an expert system, it is necessary to supply the values of various parameters. Ideally, an absolutely reliable source is available to supply an exact value for any parameter. In reality, one may have unreliable, unconfident, conflicting estimates of the value for a particular parameter. This paper is a consideration of how to represent and combine imperfect estimates.

It is assumed that the knowledge from each source takes the form of an estimate of the parameter value, paired with an associated measure of uncertainty. This uncertainty may arise from lack of confidence on the part of a human source, from some appraisal of the (un)reliability of a source, or both. Functions are proposed that combine several such value-and-uncertainty pairs to produce a resultant of the same form, that is, a resultant value and a resultant uncertainty. The proposed combination functions are most comfortably understood in terms of a parameter with values on the reals, but there is some consideration of alternative scales. The functions are developed in terms of an analogy to statistical sampling. They conform with various intuitions about the way estimates should combine.

Three related matters are worth mentioning, though space constraints preclude proper discussion. First, in addition to representing uncertain values it is also necessary to be able to compute with them, to determine both values and appropriate uncertainties for related, derived parameters in the system. (On this point, see [Garv81], and [Quin83]). Next, if there is sharp disagreement among confident and ostensibly reliable sources, one may think of reevaluating their degree of reliability. Last, there is the possibility of using information about uncertainty for deciding when it is worthwhile to seek more information that might reduce uncertainty. (Ideas related to this point are present in [Slag84] and [Duda79]).

## 2. Desirable Properties of a Solution

Intuition provides some guidelines about what it means to combine uncertain estimates. We express these guidelines as desiderata, that is, specific properties one might reasonably expect the combination functions to conform to. These desiderata provide a systematic basis for judging particular proposed combination functions. The specification of several such properties and consideration of some of their consequences form the contents of this section. Section 2.1 deals with the resultant value, and 2.2 with the resultant uncertainty.

### 2.1. The Resultant-Value Function

Resultant value depends in part on the source estimates of value, as discussed in Section 2.1.1. The associated source uncertainties should also affect the resultant value, according to the desiderata in Section 2.1.2.

#### 2.1.1. Relation to Source Values

The inputs to the resultant-value function are value-and-uncertainty pairs provided by the various sources. The output is the value that the system will use for the particular parameter. One intuition is that the resultant value should somehow reflect the central tendency of the source values that give rise to it. One way to make this precise is to mandate the mean of the source values as the resultant-value function. However, that strategy ignores the uncertainty levels of sources, thereby failing to satisfy another intuitively reasonable requirement: sources which are relatively certain should have relatively more influence. Therefore we pursue the more cautious approach of isolating and requiring some properties that underlie the notion of a mean, specifically the following.

(D1) **Range**: The resultant value should should neither exceed the largest source value nor be less than the smallest.

(D2) **Monotonicity**: If one source value increases while all others are unchanged, and while all uncertainties are unchanged, the resultant value should increase.

Statements (D1) and (D2) are less restrictive than dictating use of the mean; they do not (separately or together) require the mean, and yet the mean satisfies each of them. A potential advantage of statements like (D1) and (D2) is that although they are fully explicit, they presume existence only of ordinal relationships. Therefore in principle they can be used in conjunction with ordinal parameters, though we make no use of this here. A desirable consequence just of requiring (D1) is that if all sources give the same value that value is also the resultant value. On the other hand, an ordinal scale is too weak to express notions of symmetry and averaging.

(D3) **Symmetry**: If there are two sources and they are equally (un)certain, then the resultant value should be equidistant from the two source values.

### 2.1.2. Relation to Source Uncertainties

If one value is given with greater certainty than another, it should be counted more heavily. Since (D3) applies only to equally certain sources, we need a generalization of it for sources with unequal certainty. Further, (D3) deals only with two sources, and in general there may be several. To overcome both these limitations of (D3), we introduce (D4) which handles certainty in a quite general manner.

(D4) **Certainty**: If a source becomes more certain while its value is unchanged and while the values and certainties of all other sources are unchanged, then the resultant value becomes closer (than formerly) to the value from the source with increased certainty.

A consequence of (D3) and (D4) is that if there are two sources and the first is at least as certain as the second, then the resultant value lies at least as close to the first as to the second. One can rephrase (D4) to state that as a source becomes less certain it relinquishes its hold on the outcome. The limiting case of this is an "utterly uncertain" source, one that professes to have no knowledge whatsoever about what value to provide for the parameter in question. (D5) requires that such ignorance be ignored.

(D5) **Ignorance**: A source with utter uncertainty has no effect on the resultant.

If there are only finitely many different degrees of uncertainty, then what is meant by "utter uncertainty" is the most uncertain of these. If the degrees of uncertainty are unbounded from above, then a formal version of (D5) must make use of a limit concept, somewhat as follows: Given a set of sources, with given value estimates and uncertainties, consider the effects on resultant value and uncertainty of adding to these another source. These effects (that is, the difference in resultants with and without the additional source) approach zero in the limit as the uncertainty of the additional source is increased without bound.

It is also useful to require continuity, in its usual mathematical sense. With $K$ sources, each providing a value and a certainty, the resultant value and the resultant uncertainty are each functions of $2K$ arguments. Since a mathematical function has a fixed number of arguments, two schemas are needed, each specifying continuity of the function for each integer $K \geq 2$

(D6) **Continuity**: The resultant value and uncertainty are continuous functions of the source values and uncertainties.

### 2.2. The Resultant-Uncertainty Function

Uncertainty of individual sources surely contributes to uncertainty in the resultant. Another basis for resultant uncertainty is disagreement among sources concerning the correct value: How can a decision-maker be justifiably confident if confident, (supposedly) reliable sources provide estimates incompatible with each other? Therefore disagreement as well as individual uncertainty must play a role in determining resultant uncertainty. A third contributing factor is support, that is, weight of opinion provided by multiplicity of sources. We now examine how these three characteristics should individually and jointly affect the resultant uncertainty.

#### 2.2.1. Effects of Individual Factors

Source uncertainty should, generally speaking, have a positive effect on resultant uncertainty. The opposite effect may, however, be more appropriate in some special circumstances. Consider the case of two rather certain sources that are in strong disagreement. Presumably such apparent incompatibility warrants substantial resultant uncertainty. Now suppose that one of the sources is extremely uncertain, so uncertain that in effect it makes no contribution at all to overall knowledge. The earlier incompatibility disappears and we are left with one source, which by supposition is a rather certain one. In this way a judiciously constructed increase in individual uncertainty may reasonably be regarded as causing a decrease in resultant uncertainty. Therefore in framing a desideratum we must be careful to constrain its applicability to exclude such cases.

One strategy for excluding such cases, adopted in framing (D7) below, is to note that in the example the resultant value as well as the resultant uncertainty will have been altered. Specifically, the resultant value should move from some compromise point between the two source values to a value increasingly close to the unchanged source, as the changing source becomes very uncertain. The intuition underlying this analysis of this example has already been formulated as (D4). To capitalize on this effect, (D7) restricts attention to cases in which the resultant mean is not altered, so that if uncertainty is increased for a source on one side of the mean it also must be increased for one on the other side.

Unfortunately, even this will not suffice. For suppose that in addition to two quite certain but strongly disagreeing sources there were a third equally certain source with a value estimate midway between them. If the two sources with outlying values become utterly uncertain, disagreement can again disappear, this time with no change in the resultant value. To exclude this effect too, (D7) is fairly complex. (It is labeled "composition" to reflect that, in a limited sense, resultant uncertainty is composed of source uncertainties).

(D7) **Composition:** If the uncertainty of one or two sources increases, with no change in any of the source values or in the resultant value, and if no unchanged source has a value as close to the resultant value as that (those) of the changed source(s), then resultant uncertainty increases.

Next consider the effect of supporting opinion on resultant uncertainty. Ordinary humans with a decision to make, say about medical treatment, sometimes seek a second opinion. Presumably if the second opinion agrees with the first, even if it is no more reliable or confident, the fact of agreement makes us, the decision-maker, more confident. Indeed, one may well ask why one would ever seek a second opinion if not to gain confidence.

(D8) **Support:** If there is no disagreement about the value of the parameter (either because there is just one source or else because both/all give the same estimated value), then including an additional source with the same estimate of value reduces the resultant uncertainty (unless it is already as low as possible, in which case it is unchanged).

The notion of a supporting second opinion gives face plausibility to (D8), but the following discussion suggests a constraint on its use. Suppose that, for each source, uncertainty arises from knowing of some inherent uncertainty in the parameter value under specified circumstances. Each source itself may be absolutely reliable and confident that the parameter value can be given only with a certain degree of imprecision. Thus some estimated degree of inherent imprecision may be given by one source and then confirmed by a second source, both sources being absolutely reliable and confident. With imprecision itself interpreted as a parameter, for which there are two agreeing value estimates, the range requirement, (D1), counsels acceptance of this confirmed value of imprecision as the correct one, in violation of (D8). We nevertheless adopt (D8), but in view of the preceding argument we do so with the understanding that uncertainty is not totally inherent in what can be known about the situation, but resides to some extent in unreliability and/or unconfidence of independent individual sources.

Whereas (D7) deals with individual uncertainty and (D8) with multiplicity of sources, (D9) concerns the effect, on resultant uncertainty, of different degrees of agreement among sources. In comprehending (D9), notice that the two source values mentioned will invariably be on opposite sides of the resultant value, assuming that the monotonicity requirement, (D2), is in force.

(D9)     **Resolution:** If two source values both move closer to the resultant value without changing it, resultant uncertainty decreases.

A summary of the similarity and distinction between (D8) and (D9) is this: Uncertainty is reduced by, according to (D8), more sources that agree, and, according to (D9), more agreement among sources. A desirable consequence of (D9) in conjunction with (D2) and (D6) is the following (proof omitted). For a given set of source uncertainties, among sets of source values with a particular resultant value, resultant uncertainty is least when all the source values are the same.

### 2.2.3. Commensurability and Joint Effects

In order to compare or combine the effects of different factors, one must ensure that they are measured in the same units. Disagreement among sources has to do with differences among estimates (by the various sources) of the value of the parameter. So one expects to measure disagreement in the same units as differences along the scale of the estimated parameter. Combining disagreement and individual uncertainty will be carried out most straightforwardly if the two are mutually commensurable. Therefore it makes sense to look for a measure of uncertainty such that it too is expressible as a difference of parameter values.

One way to get such a measure of uncertainty is to let the source give a maximum and minimum value for the parameter and then relate uncertainty to the difference between the two. This notion is explored in Section 3, but some shortcomings appear in the associated combination functions. We therefore introduce an alternative proposal, in Section 4, that also yields an uncertainty value measurable in differences of parameter values. The claim that it is better in some ways than the use of maximum and minimum values is supported in terms of desiderata laid down in this section.

If the level of disagreement is commensurable with individual uncertainty, then it becomes possible to give meaningful consideration to properties that refer to both of these factors. However, it is not workable to apportion the responsibility for resultant uncertainty into a part arising from disagreement and a part arising from individual uncertainty, because of the following consequence of (D6) and (D8) (proof omitted). Given the value estimate and the uncertainty of one source, and given a second source uncertainty, there exists a possible value estimate from the second source, unequal to that from the first, such that the resultant uncertainty is less than that for either source individually.

Apportionment fails to be a useful concept because it considers only disagreement and individual uncertainty, ignoring the effect of the number of sources on resultant uncertainty. The following discussion, culminating in (D10), is directed to showing how (D10) relates to some possible intuitions about the interaction of all three of these factors: disagreement, individual uncertainty, and number of sources.

It will help to consider an extreme (admittedly unrealistic) case, one with unboundedly many sources. Suppose that one consults more and more sources and finds that both their value estimates and their uncertainties remain quite stable, as if drawn at random from some very large population of value-and-uncertainty pairs. If this process were to continue, the uncertainty of individuals would become progressively less important since we would be obtaining more and more support for each region of source values. On the other hand, the disagreement among the various values would tend to be confirmed. That is, our view of population disagreement would not tend to rise or fall. (D10) summarizes this argument. Because it makes no sense to speak of relative contributions, (D10) is obliged to require, in effect, that one of the contributions vanish in the limit. Space does not permit discussion of a desirable generalization of (D10) to deal with differing uncertainties.

(D10) **Sufficiency:** If all sources have the same uncertainty and if their estimated values are drawn (without replacement) from a fixed distribution, then as the number of sources increases without bound, the resultant uncertainty approaches a limiting value that does not depend on the source uncertainty level.

## 3. Some Candidate Solutions

Some plausible candidates for the resultant functions are evaluated against the desiderata in this section. Our own proposal receives similar treatment in section 4.

### 3.1. Weighted Mean

Use of an unweighted mean for the resultant value function is in violation of (D4) and (D5), though it does conform to (D1-3) and (D6). For a weighted mean scheme to satisfy (D4), it must always assign weights that decrease monotonically with increasing uncertainty. Such a scheme will also satisfy (D5) if it invariably assigns the weight zero to an utterly uncertain source. Note that a weighting scheme for resultant values is only a part of what is needed; it must be accompanied by a specification for resultant uncertainty.

### 3.2. Intervals with Intersection

One way to represent uncertainty is to state that a particular value may lie anywhere in some *interval*. For this purpose we specify an interval by giving its midpoint and half-length, interpreting these, respectively, as the value and uncertainty of a parameter for one source or for the resultant. (This representation requires the interval of uncertainty to be symmetric about the value estimate; to permit asymmetry would require the use of not two but three numbers). One way to combine source midpoints and half-lengths to yield a resultant midpoint and half-length is the use of intersection of the source intervals. The greatest lower bound of the sources becomes the lower bound of the resultant, and similarly for upper bounds. If these are compatible, the resultant value and uncertainty are computed from the resultant interval.

The intuitive justification for this technique is to regard each source interval as a fully reliable and confident assertion that the parameter value must lie in that interval. If a source cannot rule out any values in the domain of the parameter then that source should provide an interval that covers the entire domain. Caution here must be the province of the sources. Recklessly assuming cautious inputs, the combination technique excludes from the resultant any point excluded by any of them. If the intersection is empty, the resultant is undefined.

The intersection rule of combination obeys (D1,3,5,6) and weak forms of (D2,4,7,8). These weak forms can be formed from the original statements in each case by adding a phrase at the end like "or else remains unchanged" or "or else leaves it unchanged." In contrast, Rule (D9), resolution, not only fails to always hold, but rather always fails. That is, if two source values move closer to the resultant value without changing it, then resultant uncertainty *never* decreases (always increases if it is defined). As for (D10), recall that it essentially asserts that source uncertainties will become unimportant as the number of sources increases. There is no such effect with intersection.

### 3.3. Intervals with Cover

An alternative method, using the same interval representation as above, is to let the resultant interval be the smallest one that includes all the source intervals. Again the resultant value and uncertainty are computed from that interval, if it is finite.

A rationale for this combination method is that each source avers that the points in some interval can not be ruled out. The resultant should cover all these points and must be a single interval, according to the representation scheme. In strong violation of (D8), gaining additional sources can have the effect of increasing uncertainty, even when sources are in perfect agreement. A source that knows nothing should, by (D10), be ignored, but here it is made the determining factor in the resultant. (D5) is also violated.

### 4. A Proposal: Virtual Sampling

The combination functions for the 'virtual sampling' technique satisfy all of the desiderata introduced in Section 2. This is not to claim that they necessarily possess all imaginable desirable characteristics, nor that the statistics-motivated model on which they rest is appropriate to all types of scales. Despite these caveats, the combination functions can play these three roles: (i) a constructive proof that the listed desiderata are not mutually contradictory, (ii) a practical tool which one may regard as superior to others that fail on the desiderata, insofar as the latter seem appropriate, and (iii) a suggestion of a class of models.

### 4.1. The Sampling Idea

The analogy to sampling involves regarding the information provided by each knowledge source as corresponding to a distribution of sample means for uniformly sized samples of independently selected elements from an underlying population for that source. For each knowledge source, the value estimate is interpreted as the mean of this distribution of sample means. Uncertainty translates into a measure of the variability, specifically the standard deviation, of the distribution of sample means. This choice should facilitate the application of expert intuition, because of commensurability considerations that parallel those of in Section 2.2. Specifically, the unit of measurement for standard deviation is the same unit in which one measures differences in values of the parameter being estimated.

An important consequence of this viewpoint arises from the use of independent identically distributed random variables. The standard deviation of the sample mean decreases monotonically with increases sample size, specifically, falling as the inverse of the square root of sample size. Therefore an informal paraphrase of the viewpoint here is that relatively high certainty corresponds to relatively large sample size. Crudely but more succinctly: information diminishes uncertainty. Note, however, that this epigram need apply only to individual sources, not to the resultant.

With this interpretation, it is appropriate that the resultant value estimate be a weighted average of the source value estimates, weighted by sample size. For example, suppose that there are two knowledge sources and that they provide unequal estimates, given with unequal certainties. More specifically, let the certainties of the sources translate into standard deviations that are is 1:2 ratio. Then by the previously noted relationship between standard deviation and sample size, the corresponding sample sizes are in a 4:1 ratio. Accordingly, the resultant estimated value is one-fourth as far from the more certain estimate as from the less certain. If, say, the more certain estimate is 0 and the other is 1, then the resultant estimate is 0.2.

### 4.2. Resultant Uncertainty for Virtual Sampling

In broad outline, the computation of resultant uncertainty proceeds as follows. Expected squared distance to the resultant mean is taken with respect to all points of all the source distributions, and the result is taken to be the variance of an underlying distribution corresponding to the resultant. From this underlying population, a sample of size equal to the sum of source sample sizes is envisioned and the variance of sample means for this size of sample then becomes the resultant variance. This variance is translated back, via standard deviation, into resultant uncertainty.

What follows is a fuller account of the computation of resultant uncertainty. Let $m_i$ ($i = 1,..,K$) be the estimates of the parameter values by the $K$ available sources. Similarly, let $s_i$ be the standard deviations (corresponding to uncertainty intervals) and $v_i$ the variances. Let $m$ and $s$ be the corresponding resultants: the composite value that the system is to use for this parameter and an associated uncertainty interval.

The first step is to translate the uncertainties for the various sources into standard deviations; from these, the variances are determined: $v_i = s_i^2$. Call the lowest variance for any source $v^*$. It will be taken as the underlying variance for each source distribution. The choice to let all underlying variances be the same is not the only possible option. Its interpretation is that the differences among uncertainties will be reflected entirely by differences in sample sizes, taken up next. The sample size for each source, $n_i$, is found by dividing its variance into $v^*$; that is, $n_i = v^*/v_i$. This will make each sample size at most one ($n_i \leq 1$) and equal to 1 for the source(s) with lowest variance. (Sources with zero variance are also assigned a sample size of 1). Although non-integer sample sizes are conceptually peculiar, they fall within the domains of the relevant functions. Let $n$ denote the sum of the sample sizes: $n = \sum_i n_i$. It is now possible to state the formula for resultant mean, $m$, which also serves as resultant value estimate and will be needed for computing the resultant uncertainty.

$$m = \frac{1}{n} \sum_i n_i m_i$$

Since this average of the input $m_i$ is weighted according to the sample sizes, $n_i$, it would not be altered by a proportional shift of all the $n_i$. Notice that it would therefore be indifferent to a revised definition of $v^*$.

Turning to the resultant uncertainty, we must first compute the expected squared distance from source distributions to the resultant mean. For this purpose, consider a random variable $x$ with expectation $\mu = E(x)$ and variance $\sigma = E[(x-\mu)^2]$. Let $D(c)$ be the expected squared distance to some arbitrary point, $c$, so that

$$D(c) = E[(x-c)^2] = E(x^2) - 2c\mu + c^2.$$

For $c = \mu$, this reduces to the familiar $\sigma = E(x^2) - \mu^2$. Plugging the latter into the above expression for $D(c)$ yields, after simplification,

$$D(c) = \sigma + (\mu - c)^2.$$

Directing this result to the case of interest, let

$$u_i = v^* + (m_i - m)^2.$$

In this expression for $u_i$, the mean $\mu$ and the variance $\sigma$ from the expression for $D(c)$ have been replaced by $m_i$ and $v^*$, the mean and variance for an individual underlying source distribution. Moreover, we are interested in expected displacement from the resultant mean, so $c$ has been replaced by $m$. Weighting each source according to sample size, let

$$\bar{u} = \frac{1}{n} \sum_i n_i u_i$$

So we now have, as promised, the expected squared distance to the resultant mean with respect to all points of all the source distributions, with those distributions first subjected to the sampling interpretation. (The sampling interpretation replaces each original distribution with $n_i = v^*/v_i$ distributions with the original mean and with variance $v^*$). The result $\bar{u}$ of this process is taken to be the variance of an underlying distribution corresponding to the resultant. From this underlying population, we are to take a sample of size $n$ to get the resultant distribution which will determine resultant statistical parameters and thereby determine resultant uncertainty. Accordingly, the resultant variance is $v = \bar{u}/n$. It is necessarily smaller than the average adjusted variance, since $n$ is always larger than 1 (being a sum that must include 1). This is appropriate since the use of a sample involving more than one source entitles us to assume a more reliable sample mean. Taking the square root yields the resultant standard deviation, $s = \sqrt{v}$, which in turn determines the resultant uncertainty.

Notice the simple form of the expected squared difference, $u_i$. It is the sum of (i) an individual distribution's own variance and (ii) the squared distance from the mean of the individual distribution to the overall weighted mean for all distributions. The first term, the variance, corresponds to the uncertainty of an individual source. The second term reflects the degree of disagreement between an individual and the resultant value. After averaging, the corresponding term in $\bar{u}$ reflects the general level of group disagreement. This second term in $\bar{u}$ is a weighted average of squared differences and so may be interpreted as a variance, specifically, the variance of the means (of the sources).

### 4.3. Satisfaction of the Desiderata

The scheme just delineated satisfies all the desiderata of Section 2. To begin the justification of this claim, recall (Section 3.1) that properties (D1-6) follow from the use of a weighted mean, provided that the weights decrease with increasing uncertainty, becoming (or approaching in the limit) a weight of zero for an utterly uncertain source. These conditions are met by the current scheme.

The remaining desiderata concern resultant uncertainty. Beginning with the first of them, recall that (D7), composition, calls for uncertainty of the resultant to increase with uncertainty of what might be called its central components. Since $m$ and all the $m_i$ must not change, the $u_i$ must either all increase or all stay the same, depending on whether the originally lowest uncertainty is increased. There are two ways that a change can occur in $\bar{u}$: changes in the $u_i$ or in their relative weights $n_i/n$. Imagine that these changes take place in two stages. Changing the relative weights first will increase $\bar{u}$ because the only permissible changes are those that decrease the certainty, hence the relative weight of the lowest of the $m_i$, hence with the lowest of the $u_i$. Now take into account the changes in the $u_i$. Since any such changes must be increases, $\bar{u}$ must not decrease, and thus the overall effect is an increase in $\bar{u}$. Since the result of the changed source uncertainties is a decrease in $n$, the effect on $s$, which is the square root of $\bar{u}/n$ must be an increase. This establishes (D7), since $s$ is monotonically related to resultant uncertainty.

Next (D8), support, calls for a decrease in uncertainty as unanimous sources accumulate. That this is so follows from two considerations. First, the distance from every source value to the resultant value must be zero so that that the second term of each of the $u_i$ is zero, thus making $u_i = v^*$ for each $i$, and consequently $\bar{u} = v^*$. Therefore as sources are added, $\bar{u}$ is unchanged and $n$ increases, so that $s$, which is the square root of $\bar{u}/n$ must decrease. This establishes (D8), since $s$ is monotonically related to resultant uncertainty. For (D9), simply note that all that is involved is a reduction of $|m_i - m|$. The argument that virtual sampling satisfies (D10) and formalization of all the desiderata and proofs are left as exercises.